\documentclass{article}

\usepackage{arxiv}

\usepackage[utf8]{inputenc} 
\usepackage[T1]{fontenc}    
\usepackage{hyperref}       
\usepackage{url}            
\usepackage{booktabs}       
\usepackage{amsfonts}       
\usepackage{nicefrac}       
\usepackage{microtype}     
\usepackage{lipsum}		
\usepackage{graphicx}
\usepackage[numbers]{natbib}
\usepackage{doi}
\usepackage{enumitem}
\usepackage{multirow}

\title{\textbf{SITS-DECO: A Generative Decoder Is All You Need For Multitask Satellite Image Time Series Modelling} }

\author{ \href{https://orcid.org/0009-0003-5323-8362}{\includegraphics[scale=0.06]{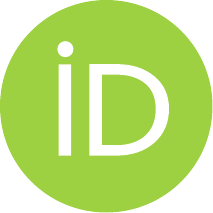}\hspace{1mm}Samuel J. Barrett}\thanks{work performed independently} \\
	LGND AI / Independent Researcher\\
    Canarias, Spain\\
	sbarrettphd@gmail.com\\
	\And
	\href{https://orcid.org/0009-0006-2021-7145}{\includegraphics[scale=0.06]{orcid.pdf}\hspace{1mm}Docko Sow}\textsuperscript{$\ast$} \\
	Tolbi / Independent Researcher\\
	Dakar, Senegal\\
	dockosow97@gmail.com}

\renewcommand{\shorttitle}{SITS-DECO}

\hypersetup{
pdftitle={SITS-DECO: Generative Sequence-to-Sequence is All You Need for Satellite Image Time Series Modelling},
pdfsubject={cs.LG},
pdfauthor={Samuel J. Barrett, Docko Sow},
pdfkeywords={Generative AI, Earth Observation},
}

\begin{document}
\maketitle

\begin{abstract}
Earth Observation (EO) Foundation Modelling (FM) holds great promise for simplifying and improving the use of EO data for diverse real-world tasks. However, most existing models require additional adaptation before they can be used and are structured rigidly around particular data sources or training approaches. To address this, we take inspiration from large language models, where diverse tasks, both pre-training and downstream, are implicitly captured through next-token prediction over unified token sequences, leveraging the structure and diversity of the training data.

We introduce SITS-DECO (Satellite Image Time Series-DECoder Only), a proof-of-concept generative model that applies this unified-sequence framing to EO data. Using a simple GPT-style decoder-only architecture, and demonstrate its ability to perform useful EO tasks (pixel-wise, multi-temporal, multi-modal crop-type classification) in a purely generative framework. Through symbolic prompting, we show that the model can perform multiple supervised and self-supervised tasks within a single unified architecture—without task- or modality-specific adaptation. Despite its simplicity and lack of spatial context, SITS-DECO outperforms much larger EO foundation models on crop-type classification (PASTIS-R) demonstrating that dense temporal sequence modelling is a critical missing ingredient in the current paradigm.

This work exemplifies a data-centric modelling paradigm in which capability arises from the diversity and structure of the training data rather than from architectural complexity. SITS-DECO provides a lightweight, practical route to multi-modal, multi-task EO modelling, and a conceptual bridge toward future generative EO foundation models.
\end{abstract}

\section{Introduction}
Traditional machine learning models in sequential domains typically learn mappings from sequences to discrete values or classes—such as predicting sentiment from text or land-cover type from satellite image time series. In natural language processing, the sequence-to-sequence paradigm became a major and influential framework, enabling tasks such as translation and summarisation to be expressed within a single model architecture rather than as fixed-label classification problems. The next true conceptual shift came with decoder-only generative models such as GPT \cite{radford2018improving}, which collapse the distinction between input and output entirely: all tokens, whether contextual or predictive, inhabit the same representational space. Through next-token prediction over unified sequences, such models implicitly acquire a broad range of capabilities without task-specific supervision or architectural modification.

This unified-sequence formulation is not only conceptually elegant but also practically transformative. It provides the foundation that makes large-scale pre-training effective: once diverse tasks are expressed within a shared token space, scaling data and model capacity naturally yields emergent capability. In contrast, models such as BERT \cite{devlin2018bert} represent a parallel scaling paradigm—one centred on self-supervised representation learning rather than generative prediction. This approach, pre-training large models on unlabelled data and then adapting them to downstream tasks, has become the dominant recipe for foundation models across domains.

Earth Observation foundation models largely follow this trajectory, adopting large-scale pre-training and transfer learning from NLP and vision. Yet these models retain the representational framing of BERT rather than the generative unification of GPT, relying on task-specific heads, fine-tuning, or linear probes for adaptation. Extending the unified-sequence, decoder-only approach directly to EO data offers a path toward true multitask generality: a model that learns from and reasons about satellite observations entirely within a shared token space, without language intermediaries or post-hoc architectural modifications.

A complementary route toward adopting this generative paradigm in Earth Observation involves integrating EO data directly into large multimodal foundation models, alongside natural images, language, and other modalities. This approach can confer task flexibility through shared multimodal representations but introduces high computational cost and indirect supervision via language. In parallel, an alternative pathway is to apply the unified-sequence, decoder-only principle within the EO domain itself, developing models that achieve similar generality without dependence on full language scaffolding.

Across the EO literature, pre-training remains fragmented across supervised, self-supervised, and contrastive paradigms, each optimised for particular data types or downstream tasks. This reflects the absence of a unifying modelling framework rather than a limitation of the data itself. By framing EO modelling within a generative, decoder-only paradigm, these approaches can be brought under a single formulation: different tasks become alternative ways of conditioning or predicting within the same token space. This not only clarifies the role of self-supervision but expands it—opening the door to flexible combinations of supervised, self-supervised, and generative objectives within one consistent architecture.

\subsection{Proposed Approach}
Building on this conceptual framing, we present SITS-DECO, a flexible proof-of-concept model that represents diverse Earth Observation (EO) tasks, both pretext and downstream, as unified token sequences combining continuous data and symbolic elements. Inspired by the unified-sequence paradigm exemplified by GPT-style models, SITS-DECO performs all tasks within a single decoder-only architecture, eliminating the need for task-specific encoders or output heads. By explicitly encoding tasks as symbolic tokens alongside data tokens, our model generalises across multiple EO tasks and sensor modalities. Critically, this hybrid continuous–symbolic token sequence approach supports generative pre-training, enabling flexible multi-task inference through simple task prompting without architectural modifications, specialised heads, or full language modelling traditionally required. The use of continuous tokens is a deliberate simplification chosen to support immediate practical utility along with exploration of pre-training techniques, but full GPT compatibility would require discretisation of the continuous tokens.

\subsection{Multi-Temporal Modelling as a Demonstrator and Motivator}
We specifically demonstrate this concept using pixel level multi-temporal satellite image time series (SITS), motivated by three key reasons:
\begin{enumerate}
\item Capability gap: Multi-temporal EO data is critically important for numerous real-world applications but remains relatively poorly supported by current EO foundational modelling approaches, which primarily focus on spatial and multi-modal tasks on single or few time step images (although some recent embedding models are stronger on multi-temporal modelling).
\item Ideal fit for sequence modelling: Transformer architectures naturally excel at sequence-based modelling. Multi-temporal EO data is inherently sequential, making it an ideal test bed to demonstrate and evaluate our token sequence approach.
\item Computational and architectural simplicity: Modelling pixel-level multi-temporal data allows us to avoid the computational and architectural complexity typically associated with spatial image tokenization (e.g. ViTs), ensuring our approach remains lightweight, flexible and accessible.
\end{enumerate}
Thus, while improved multi-temporal modelling itself is valuable, our central contribution is demonstrating the power and flexibility of representing EO tasks as unified token sequences within a generative, decoder-only transformer framework as a fundamental EO modelling paradigm.

\subsection{Challenges and Limitations of Existing Approaches}
There are currently two main families of approach to EO foundational modelling: representation-centric models, and instruction-centric models.

Representation-centric EO FMs (examples: Prithvi \cite{prithvi2023}, Clay \cite{clay_spec_2024}, TerraMind \cite{terramind2025}, Presto \cite{presto2023}), are mostly used for downstream tasks by either attaching task specific heads and fine-tuning, or modelling on embeddings/frozen model weights (e.g. linear probing). They are not directly usable for downstream tasks without additional adaptation. Further, their architectures are often built around the modalities and training objectives used, so any additional modality or pre-training task requires architectural changes. Lastly, these models tend to perform poorly on tasks requiring dense multi-temporal data and understanding, even when combined with a bolt-on temporal encoder (see related work). A special subset of representation-centric models are EO embedding models which are specifically designed and trained to be used to generate embeddings for downstream retrieval or modelling. So far few EO FMs were primarily designed as embedding models with the recent exceptions of Tessera \cite{tessera2025} and Google’s AEF \cite{aef2025}.

Instruction-centric (prompt-to-task) models (examples: EarthDial \cite{earthdial2024}, Falcon \cite{falcon2025}, EarthGPT \cite{earthpt2023}), mostly CLIP-like or (multi-modal) GPT style approaches, use language to define the task and are thus promptable to tasks the model has been trained on, and can be easily adapted to new tasks by incorporating new instruction tuning examples (EO-data + task as text + task output). However, incorporating language to gain this flexibility is computationally expensive. Furthermore, few have explored training on dense multi-temporal EO data, and effective multi-modal and multi-temporal EO data pre-training in a compatible token sequence framework is not well explored.

\subsection{Contributions}
\begin{enumerate}
\item \textbf{Core conceptual contribution}\\
We introduce and deeply explore a general-purpose Earth Observation modelling paradigm inspired by GPT-style transformer architectures and modelling approaches, explicitly encoding diverse EO tasks as sequences of continuous data tokens (reflectances, backscatter values, lat-lon) and symbolic tokens (task instructions, classes).
\item \textbf{Practical demonstration on multi-temporal EO data}\\
We demonstrate the effectiveness of this framework on multi-modal, multi-temporal satellite image time series (SITS) modelling, showcasing strong performance on multiple framings of a classic temporal challenge: crop-type classification. In particular, we show strong multi-modal and multi-temporal performance, the ability to incorporate contextual metadata (e.g. spatial identifiers or coordinates), and robustness to irregular, noisy, and unaligned observations across modalities.
\item \textbf{Promptability via symbolic task tokens}\\
The unified token-sequence design allows the model to perform multiple tasks learned during training through simple symbolic prompting—without fine-tuning, task-specific heads, or language input. Once trained, the same model can be prompted to execute any of its learned tasks directly, demonstrating true multi-task inference within a single framework.
\item \textbf{Zero-adaptation extensibility}\\
Adding new tasks or modalities requires no architectural modification. Because all tasks are represented as token sequences, new capabilities can be introduced purely through data by including new task examples during training. This data-driven extensibility supports flexible task addition, mixed-task pre-training, and new approaches to domain adaptation.
\item \textbf{Insights on reduced preprocessing complexity and simple multi-modal fusion}\\
Transformers naturally handle irregular time series, sparse observations, multi-modal data and minimal pre-processing, significantly simplifying traditional EO pipelines.
\item Collectively, these contributions, together with the lightweight architecture (no language, no spatial operations, pixel-time-series only), offer both immediate practical usability and a clear conceptual bridge toward future, fully generative EO foundation models.
\end{enumerate}

\subsection{Motivating Use Case}
While many of the above contributions have potential broad applicability, a core original motivation for this work concerns agricultural monitoring. Agriculture is widely recognized as a critical application domain for EO given its wide geographic extent (roughly 38\% of the Earth’s land surface \cite{zabel2019_global_cropland_impacts}), and food security and economic importance. Within agricultural monitoring, EO data is widely used for food security monitoring, precision farming, sustainability and environmental monitoring.

Concerning the above described contributions: Agriculture is arguably the most dynamic large scale earth surface land cover/use type, meaning that \textbf{multi-temporal} data is critical for many specific agricultural monitoring tasks. The domain as a whole also includes a wide range of EO data derived tasks (each contributing to important applications mentioned above), hence the potential value on \textbf{promptability and flexible extensibility} to support tasks including: crop mapping, yield estimation/forecasting, management practice and event (e.g. harvest) detection, crop performance monitoring, crop damage/disease detection, soil moisture estimation and more. Further, many of these tasks benefit from multiple modalities, or suffer from the challenges of cloudy observations in time series of optical imagery. Therefore there is value in simplifying the \textbf{data pre-processing and data fusion} methods often used to address these challenges. As a widely studied and benchmarked representative task, we focus specifically on crop type classification/segmentation in this work.

While agriculture is our original motivator and demonstrator domain, the contributions of this approach are generally and widely applicable to a range of other multi-temporal EO modelling scenarios, such as forestry monitoring, land cover/use change detection. Further, we believe these contributions with adaptation to be widely applicable beyond just application domains where multi-temporal data is important.

\subsection{Summary of Contributions}
In summary, this work introduces a unified, symbolic- and continuous-token sequence decoder-only transformer framework, enabling flexible generative pre-training and zero-adaptation inference across diverse Earth Observation tasks, sensors, and modalities. Our hybrid token approach simplifies multi-modal and multi-temporal modelling, greatly reducing preprocessing complexity and improving practical usability compared to existing EO foundational models.

\subsection{Overview of Paper Structure}
\begin{itemize}
    \item Section 2 reviews related work in transformer based EO modelling, foundation modelling and multi-temporal modelling.
    \item Section 3 describes the benchmark dataset used in this work.
    \item Section 4 describes the proposed architecture, data processing, data representations, task definitions and training setup along with the description of experiments run.
    \item Section 5 reports the experimental results.
    \item Section 6 discusses the key findings and limitations.
    \item Sections 7 and 8 discuss possible future directions and conclusions.
\end{itemize}

\section{Related Work}
\subsection{Transformers in NLP, Vision and Multi-Modal Modelling}
Transformer neural networks have become the foundational technology of AI development in recent years. The seminal paper \cite{vaswani2017attention} concerned machine translation and used an encoder-decoder architecture with “multi-headed attention” instead of recurrent (or convolutional) layers to process sequences of language tokens. These attention-based models are effective in natural language processing because they excel at modelling long-range dependencies primarily due to their attention mechanism explicitly relating different positions in the input. Further, in contrast to recurrent models (the previous paradigm in NLP) which process sequences sequentially, transformers are fully parallelisable (not recurrent) and so can be effectively scaled on GPUs. Two distinct paradigms using transformers for natural language processing quickly arose: 1. Encoder-only models like BERT \cite{devlin2018bert} and 2. Decoder-only models like GPT-1/2/3 \cite{radford2018improving}. A key innovation in GPT-2 \cite{radford2019language} and GPT-3 \cite{brown2020language} was their framing of various NLP tasks as prompt-based unified token-sequence tasks, enabling generalisation across diverse tasks without requiring architectural modifications. In contrast, BERT-style models typically require adding and fine-tuning specialized output heads.

While transformers were originally designed for language processing tasks, their flexible attention-based architecture also lent itself to other domains. In computer vision (CV), Vision Transformers (ViTs) emerged as an alternative to the dominant convolutional neural networks (CNNs). Rather than operating on pixels (which would be computationally prohibitive due to the quadratic memory and compute requirements for sequence length), ViTs operate on flattened sequences of patches, each embedded as a token. ViTs have been shown to be powerful when trained on sufficiently large datasets \cite{dosovitskiy2020image} and excel, as in language, at global relationships in contrast to CNNs which primarily model local scale spatial relationships.

Transformers have also been widely used in multi-modal (mainly but not limited to text + images) modelling. An early important paradigm was the use of bi-tower (an image encoder and a text encoder) architectures like CLIP \cite{radford2021clip} trained to align representations via contrastive learning. More recently there’s a trend towards unifying all modalities in a single unified “everything’s a token” architectures. Frontier models like Google’s Gemini 2.0 \cite{gemini2023} and OpenAI’s GPT-4o (“o” for “omni” \cite{openai2024gpt4o}) can process sequences containing tokens representing: text, images, audio and video.

\subsection{Foundation Modelling}
In 2021, the Stanford Center for Research on Foundational Models (CRFM) published \cite{bommasani2021foundationmodels} a comprehensive analysis of a broad paradigm shift in AI which they termed foundation modelling. They described the trend of increasingly large models, pre-trained at scale and adapted to a wide range of downstream tasks which has become standard practice in NLP and CV. The core value is that massive (usually self-supervised) pre-training enables the learning of general representations which can be used as a foundation for downstream modelling tasks without the data or compute requirements necessary for training from scratch.

\subsection{EO FMs}
Earth Observation represents a potentially powerful application domain for foundation modelling. There are huge (petabyte scale - \cite{geo2023copernicus_dataspace}) archives of imagery from diverse modalities suitable for pre-training, though for many downstream applications, relatively limited labelled datasets. Therefore, pre-training in general and large scale foundational modelling efforts in particular are progressing rapidly. The recent history of EO Foundational modelling is summarised in a recent benchmark paper \cite{pangaea2024}. Earth Observation is a broad domain, and that breadth is expressed in many different approaches to foundation modelling emphasising various different considerations besides simple architectural modifications. We characterise existing EO Foundational Models along ten critical axes that impact their practical utility and flexibility:
\begin{enumerate}[label=\arabic*.]
    \item The pre-training objective, most commonly either:
        \begin{enumerate}[label=\alph*)]
            \item Masked auto encoding + related: e.g. SatMAE \cite{satmae2022}, Prithvi \cite{prithvi2024}, Clay \cite{clay_spec_2024} and Google AEF \cite{aef2025}
            \item Contrastive learning + related: e.g. TerraMind \cite{terramind2025}, Galileo \cite{galileo2025}, RemoteCLIP \cite{remoteclip2023} and Tessera \cite{tessera2025}
        \end{enumerate}
    \item Which and how many (EO) modalities are included (as inputs):
        \begin{enumerate}[label=\alph*)]
            \item Optical RGB only, e.g. RemoteCLIP \cite{remoteclip2023}
            \item Optical multi-spectral only, e.g. SatMAE \cite{satmae2022}
            \item Optical + Radar, e.g. CROMA \cite{croma2023}
            \item Optical + Radar + other, e.g. Galileo \cite{galileo2025} and TerraMind \cite{terramind2025}
        \end{enumerate}
    \item Whether spatial modelling is included: the vast majority do, though some models are pixel based (i.e. spectral-temporal, e.g. PRESTO \cite{presto2023}, Tessera \cite{tessera2025})
    \item Whether text is included as a modality:
        \begin{enumerate}[label=\alph*)]
            \item For prompting, see below.
            \item For captioning: e.g. TerraMind \cite{terramind2025}
            \item For supervision: e.g. Google AEF \cite{aef2025}
        \end{enumerate}
    \item Whether they are promptable:
        \begin{enumerate}[label=\alph*)]
            \item Via text - instruction tuned LLMs: e.g. SkyEyeGPT \cite{skyeyegpt2024} and GeoChat \cite{geochat2023}
            \item Via text - CLIP EO-text alignment: e.g. RemoteCLIP \cite{remoteclip2023}
            \item Via symbolic tokens: e.g. TerraMind \cite{terramind2025}
        \end{enumerate}
    \item Whether they are generative:
        \begin{enumerate}[label=\alph*)]
            \item Time series forecasting: e.g. EarthPT \cite{earthpt2023}
            \item Image/map generation: e.g. TerraMind \cite{terramind2025}
            \item Text generation: e.g. Falcon \cite{falcon2025} and EarthDial \cite{earthdial2024}
        \end{enumerate}
    \item How multiple modalities are unified (see review paper: \cite{lane_karimzadeh_2025_genealogy_ms_fms_rs}), for example:
        \begin{enumerate}[label=\alph*)]
            \item Separate tower/encoder per modality with no feature fusion, aligned via contrastive loss, e.g. RemoteCLIP \cite{remoteclip2023}
            \item Separate encoders per modality with feature fusion, e.g. CROMA \cite{croma2023} and TerraMind \cite{terramind2025}
            \item Shared encoder architectures, e.g. USat \cite{usat2023} and Galileo \cite{galileo2025}
        \end{enumerate}
    \item If/how multi-temporal imagery is handled (see \cite{galileo2025} for review):
        \begin{enumerate}[label=\alph*)]
            \item Mono-temporal only: e.g. SatMAE \cite{satmae2022} and CROMA \cite{croma2023}
            \item Few-observation (e.g. quarterly): e.g. Prithvi \cite{prithvi2024}
            \item Many-observation (e.g. monthly): e.g. PRESTO \cite{presto2023} and Galileo \cite{galileo2025}
            \item Dense (e.g. sub-weekly): e.g. EarthPT \cite{earthpt2023} and Tessera \cite{tessera2025}
        \end{enumerate}
    \item Whether the goal is representation learning or direct general purpose promptable utility (representation-centric vs instruction-centric):
        \begin{enumerate}[label=\alph*)]
            \item Representation-centric: e.g. Prithvi \cite{prithvi2024} and Clay \cite{clay_spec_2024}
            \item Instruction-centric: see 5/5a above
        \end{enumerate}
    \item Whether they produce outputs at patch or pixel level:
        \begin{enumerate}[label=\alph*)]
            \item Patch: e.g. EarthGPT \cite{earthgpt2024}
            \item Pixel: e.g. TerraMind \cite{terramind2025} and Falcon \cite{falcon2025}
        \end{enumerate}  
\end{enumerate}

Note, the examples above should be seen as instructive. This list is not an exhaustive analysis of the capabilities of any particular EO foundation model.

Our work on these axes:
\begin{enumerate}[label=\arabic*.]
    \item Different pre-training objectives have different strengths. We aim to develop a modelling framework compatible with diverse objectives instead of being built around a single one.
    \item Most models are architected specifically around the modalities included. We aim for a general purpose framework, theoretically compatible with any arbitrary modality (ala GPT 2->4o “everything’s a token”)
    \item We deliberately omit explicit spatial modelling in this study to isolate and investigate temporal modelling, token representation and pre-training objectives. This also keeps the model light-weight for downstream uses where spatial modelling is less critical. However, future work could integrate spatial patches within the same flexible framework.
    \item We choose not to include text as a modality because of the significant computational overhead, though we choose an architecture and modelling approach (token sequences and symbolic task tokens) fundamentally compatible with modern large language models to enable scaling and incorporation of language in future.
    \item We adopt symbolic token prompting to ensure direct usability, eliminating the need for additional training steps (fine-tuning, linear probing) after pre-training, significantly reducing downstream deployment complexity.
    \item We choose a generative model both for fundamental compatibility with LLMs, and to enable single and multi-task flexibility.
    \item We choose a single decoder only architecture for simplicity and compatibility with LLMs.
    \item We choose to focus on dense “many-observation” multi-temporal data to address the significant limitations on multi-temporal task performance of most current EO FMs.
    \item Our goal is direct promptable utility rather than representation learning to remove the need to adapt the model to downstream tasks post-training.
    \item Our model being non-spatial and relatively light-weight can be used per-pixel.
\end{enumerate}
Additionally, we prioritize architectural simplicity, pushing as many questions as possible from the architectural choices space to the data representation space.

\subsection{Multi-temporal EO FMs}
There’s a rich history of various modelling approaches to multi-temporal data (with or without spatial modelling) in Earth Observation, and within the task of crop classification which is the example application task used in this paper. Classic ML methods such as SVM (e.g. \cite{devadas2012svm}), Random Forest (e.g. \cite{blickensdorfer2022rf}), and gradient boosting machines (e.g. \cite{goldberg2021_s2_cropmaps_israel}) have been widely used. Within deep learning, the time dimension has been modelled variously with (temporal) convolutions (e.g. \cite{pelletier2019tempcnn}), recurrence (e.g. \cite{russwurm2018rnn}) and attention (e.g. \cite{russwurm2019_self_attention_sat_ts} and \cite{ltae2020}).

Multi-temporal foundation models are less common than those which are purely spatial. One approach to using atemporal EO FMs for multi-temporal tasks, used in the PANGAEA benchmark of many EO FMs \cite{pangaea2024}, is to compute spatial embeddings on each image, then combine the resulting temporal embedding stack using a dedicated temporal encoder or linear mapping. Several EO FMs are trained on multi-temporal data, though only on relatively few images. For example Prithvi \cite{prithvi2024} handles multiple images during patch embedding (trained on 3 time steps) and SatlasNet uses temporal max pooling (of up to 8 Sentinel 2 images) to combine the spatial embeddings \cite{satlaspretrain2022}.

There are still fewer models which incorporate longer or more dense multi-temporal data. Presto (pixel time series only) \cite{presto2023} and Galileo (spatio-temporal) \cite{galileo2025} both use 12 monthly imagery composites and are applied to downstream tasks by using only the pre-trained encoder (the pre-training of both uses a paired encoder-decoder). One clear outlier is EarthPT \cite{earthpt2023}, a GPT decoder only pixel time series model which is trained on 5-day cadence data covering up to 7 years. It is has been applied to downstream tasks via linear probing (presumably on the representation from the final time step) \cite{earthpt_linearprobe_2025}. Further, the recent embedding-focused models Tessera and Google’s AEF are both trained on dense time series (all or most of available observations in a given period).

Also of note is Earth Dial \cite{earthdial2024} which can incorporate multi-temporal imagery but only for scene classification (not pixel level semantic segmentation), so is less related to the above mentioned models.

\section{Benchmark and Evaluation Dataset}
For this work we use the PASTIS-R \cite{pastisr2021} crop segmentation dataset (19 crop classes + background) for the following reasons:
\begin{enumerate}
    \item It is a multi-modal and multi-temporal dataset including data from Sentinel 1 and Sentinel 2.
    \item It does not extensively pre-process the EO data, enabling testing minimal pre-processing and data fusion techniques.
    \item It is widely used for benchmarking EO FMs and is included in the PANGAEA benchmark \cite{pangaea2024}.
\end{enumerate}

The dataset is designed for panoptic and semantic segmentation of crops, though we only use it for semantic segmentation (crop classification per pixel) in this work. It contains 19 crops across 4 Sentinel tiles in France. The labels and field geometries are derived from French subsidy (GSAA/LPIS) data. The data is 10m resolution with 2,433 patches of 128x128 pixels. The dataset specifically does not pre-address key issues like cloud corrupted observations, class imbalance and irregular time-series, allowing those issues to be addressed by the user.

PASTIS-R contains 5 folds, designed to include representative distributions of crops and uses non-adjacent patches to avoid spatial autocorrelation. We note that the original PASTIS-R evaluation protocol differs from that used in PANGAEA. In our results we use both approaches (original leave-one-out cross validation [5 combinations of train on 3, val on 1, test on 1], and PANGAEA’s train on folds 1, 2 and 3, validate on 4 and test on 5 [the first fold combination from the original protocol]) so we can compare to both PANGAEA benchmark results and other models benchmarked on the original protocol. For comparison to published benchmark results, we focus on reporting mIoU, but also mention overall accuracy and mean class accuracy.

\section{Methods}
\subsection{Model Architecture}
We adopt a minimalist decoder-only transformer architecture directly inspired by GPT-2 (\cite{radford2019language}, \cite{vaswani2017attention}) and the recent EarthPT model \cite{earthpt2023}, which successfully demonstrated GPT-style modelling specifically for pixel-level satellite time series data. Similar to EarthPT, our architecture emphasizes simplicity, flexibility, and direct applicability to pixel-level sequences.
Specifically, our model consists of:
\begin{itemize}
    \item 2 transformer decoder blocks, each employing multi-headed self-attention with 16 attention heads per block.
    \item 128-dimensional input embeddings for both symbolic (discrete) tokens and continuous sensor data tokens.
    \item Standard learned positional encoding: day since start (2018‑09‑16) for data tokens and positional index for other tokens.
    \item A single unified linear projection head generating predictions across all token types (categorical and continuous), with prediction types identified by preceding symbolic task tokens.
\end{itemize}

This deliberate architectural simplicity is not a constraint but a design principle. By shifting task, modality, and data complexity into the token representation itself, we ensure that the model architecture remains generic and reusable. In this way, architectural minimalism directly enables the data-centric flexibility that underpins our framework.

Unlike EarthPT, we do not incorporate explicit date-hint embeddings, further simplifying the architecture. We intentionally avoid encoder-decoder structures or modality-specific adapters, emphasizing instead the model’s capability to generalise to diverse EO tasks through symbolic prompting alone.

\subsection{Pre-processing}
The Sentinel 1 backscatter and Sentinel 2 reflectance values from the PASTIS-R dataset are converted to floating point values and scaled by 1/10 and 1/10,000 respectively which makes their ranges comparable and roughly -1 to 1 (this is deliberately done in lieu of the added pipeline complexity of full normalisation). For each observation we add a “days since start of PASTIS time-frame” (2018-09-16) channel in position 0 (used for positional encoding only) and metadata including the patch centroid lat-lon and the tile ID are made available as tokens for associated task representations. The crop label data is one-hot encoded but otherwise unaltered. As the model operates on pixel time series, and each sample is a task based on a single pixel, we flatten the patches and shuffle/sample pixels from the flattened representation. To ensure mixing of samples from different patches within a batch, we load a number of patches into memory (default: 64) and load and shuffle pixel samples across the loaded patches. Further this is done across several parallel workers (default 8). As each batch is fed by all workers, each worker having a large buffer of patches to sample from, each batch contains diverse samples. This setup is necessary because the entire dataset is too large to store in RAM during training.

To reduce the length of Sentinel 2 time series to speed up model training, a maximum number of Sentinel 2 time steps can be chosen (default 30). A very simple cloud probability score is calculated (NDSI) and the least likely to be cloudy observations are selected. No compositing, interpolation, time-series alignment or robust cloud masking is performed (i.e. the resulting Sentinel 2 data is highly irregular). In case of trying to reconstruct Sentinel 2 data (see below), it is necessary to prevent the model trying to reconstruct cloudy observations. Therefore we construct a sample weighting using the above simple cloud probability score (NDSI > -0.3) to create a binary sample time step weight (0 for maybe cloud, 1 for probably clear).

PASTIS-R provides Sentinel 1 ascending and descending orbit data separately. Rather than concatenating these in the time/sequence dimension, to reduce the overall sequence lengths, we concatenate them in the channel dimension. While the Sentinel 1 revisit and cadence is mostly constant, there are some anomalies. Therefore we match the closest date pairs of ascending and descending orbits for channel dimension concatenation in the same temporal position and simply use the day of year from the descending orbit. This introduces some noise and is a design choice for short term simplicity and training efficiency.

Observation level dropout augmentation is also optionally applied (default: True) to both the Sentinel 2 and Sentinel 1 data. In this case, a proportion of samples to augment is defined, and for each “to-be-augmented” sample, a maximum and minimum observation dropout rate is defined. By default, 50\% of samples have this augmentation applied, and between 0 and 95\% of observations are dropped out per sample. In this case, the samples are removed (not just set to 0) as sequence length consistency is handled after task representation with simple padding.

\subsection{Task Representation}
All tasks are constructed from sequences of data and symbolic tokens. To make loss calculation and output handling simple in the training and inference implementations, the output is symbolically split in the channel dimension into a section for continuous values (containing channel dim sections for each sensor/modality) and a discrete section (containing one hot encoded representations of symbolic task and discrete data tokens). The token types are:
\begin{enumerate}[label=\Alph*)]
    \item Continuous data tokens (Sentinel 1, Sentinel 2, lat-lon)
    \item Discrete/categorical data tokens (crop type, tile ID) - one-hot encoded
    \item Symbolic task tokens - one-hot encoded
\end{enumerate}
Instead of defining tasks based on what has come before and what should be predicted after, we generalise model behaviour to: “given whatever is in the sequence before, now predict token type X”. Therefore these task tokens are simple identifiers which during training identify the type of following token (e.g. Sentinel 2 data), and during inference identify the type of token to generate (e.g. Crop class). These tokens are also one-hot encoded.

\subsection{Defining Tasks}
To enable easy configuration of different task sequences, they can be defined in yaml configuration files which include a list of token types and parameters from which to construct a sequence. 

For example, if we wanted to do Sentinel 2 based crop type classification, symbolically:
{\texttt S2 -> Crop}

We would express that in the configuration (simplified form) as follows:

{\texttt[symbolic task token: S2, S2 data tokens, symbolic task token: crop, crop categorical token, end of sequence token]}

Which yields a sequence like (where <> indicates a discrete token):

{\texttt[<task: S2>, S2t0, S2t1, … S2tn, <task: CROP>, <crop\_class>, <EOS>]}

As the model is trained with teacher forcing, each token is predicted based on all previous tokens. Tasks can therefore be chained together, and by default every categorical data token is included in the loss calculation, so tasks with multiple categorical data tokens implicitly contain multiple tasks.

For example the task:

{\texttt S2 + tile ID -> Crop}

Is implicitly:

{\texttt S2 -> tile ID -> Crop}

Which means the model will simultaneously learn both to predict tile ID based on Sentinel 2, AND predict crop based on Sentinel 2 AND tile ID.

\subsection{Self-Supervised Tasks}
We principally implement an instance discrimination style self-supervised task inspired by ALISE \cite{alise2024}, though we intend for the modelling framework to support more diverse tasks, and we also experimentally implement regression tasks (see below) which support a masked reconstruction self supervised task.

For instance discrimination, we take a single simple approach, though again, the modelling framework can support additional strategies simply via how token sequence samples are constructed. We train the model to distinguish if pairs of Sentinel 2 and Sentinel 1 sequences are from the same pixel or not. For this task type, for a given pixel, we take the Sentinel 2 time series data, and with a random coin flip (p=0.5) either use the Sentinel 1 data from the same pixel, or from another pixel stored in a buffer of recently processed pixels (diverse, from multiple tiles, patches and training folds). We then append a discrimination task token and a token representing either “match” or “mismatch” depending on the coin flip. Therefore we generate 2 kinds of sequences:

Match (all pixel a):

{\texttt [<task: S2>, aS2t0, aS2t1, … aS2tn, <task: S1>, aS1t0, aS1t1, … aS1tn, <task: discrimination>, <MATCH>, <EOS>]}

Mismatch (Sentinel 2 pixel a, Sentinel 1 pixel b):

{\texttt [<task: S2>, aS2t0, aS2t1, … aS2tn, <task: S1>, bS1t0, bS1t1, … bS1tn, <task: discrimination>, <MISMATCH> , <EOS>]}

This approach can easily be extended to other combinations and views (e.g. sub-views of a single modality) simply by adding view generation functionality to the token sequence generator. 

The above task is fully self-supervised, but the same discrimination approach can be used with labels. For example, given a Sentinel 2 time series and a crop label which may or may not match the correct label for the pixel, generate match/mismatch. The value of this kind of task is not investigated in this work, but may be useful for downstream tasks like validating reported labels (e.g. farmer reported crops in CAP subsidy checks) as the problem of “does this data match the reported crop” is a more constrained problem then accurately predicting the actual crop from the data.

\subsection{Regression Tasks}
The core focus of this work is for classification based tasks (even self-supervised ones). However, the concept and code both support regression targets as well - that is predicting continuous value tokens not just discrete ones. This capability is described separately because it is not as deeply explored in this work, but the possibility and potential value align closely with the motivation and key contributions. This capability can be used both as a downstream task (e.g. forecasting or cloud gap imputation) or for self-supervised learning via various masked reconstruction strategies.

Given training is via teacher forcing - i.e. “predict the next token”, the only change needed to implement regression tasks given the way task sequences are constructed is to calculate a loss on the continuous tokens and combine it with the categorical token loss. To achieve this, our data pipeline keeps track of which time steps are continuous and which are discrete tokens as sample time step weight vectors which can be applied in the training loop. As mentioned before, if we try to reconstruct Sentinel 2 data, it is necessary to discourage the model from predicting clouds, so any possibly cloudy observations are given a regression time step weight of 0.

When enabled, regression tasks (i.e. prediction of continuous data tokens) use MSE or Huber loss and are combined by a configurable weighted (typically 0.01) sum with the discrete token cross entropy loss.

Satellite image time series data are highly temporally auto-correlated. For this reason, simply predicting the next observation in a time series is not necessarily a very difficult task and might not provide a useful training signal (although it seems sufficient in EarthPT \cite{earthpt2023}). Therefore, to encourage better representation learning and more difficult sensor to sensor translation tasks, we also provide the ability to mask out time steps from a sensor in the transformer block attention masks. Possible schemes are random (some random proportion of time steps are masked), patch (continuous patches of some length are masked), or complete (all observations are masked). The complete option is primarily useful for teaching single-shot cross sensor translation.

It should be noted that the Sentinel 2 cloud filtering approach and Sentinel 1 \& 2 observation level dropout make the time-series irregular. We address this in our limited regression experiments by turning off Sentinel 2 cloud filtering and observation level dropout to ensure the time-series are (mostly) internally regularly spaced. Another approach would be to prompt the model with the date to reconstruct as done in EarthPT \cite{earthpt2023}.

\subsection{Training Setup}
As described above (Regression Tasks), our dataloader keeps track of which tokens are continuous (if using regression) and which are discrete (for classification) via a pair of binary loss weights vectors, one for regression one for classification. Symbolic tokens are 0 in both vectors as they are never generated, only prompted. These weights vectors are multiplied by the regression and classification losses per time-step per sample respectively to effectively apply the appropriate loss to each time-step. For classification we use categorical cross entropy loss, and for regression we use MSE or Huber loss (after EarthPT \cite{earthpt2023}). The mean loss is calculated separately for regression and classification then are combined via a weighting parameter (see above).

Teacher forcing is used with a causal mask such that the model predicts each output token given all the preceding tokens as context. Augmentation via observation drop out and masking is described in the pre-processing and regression task sections above.

\subsection{Experiment Design and Evaluation}
\begin{figure}
	\centering
    \includegraphics[width=0.8\linewidth]{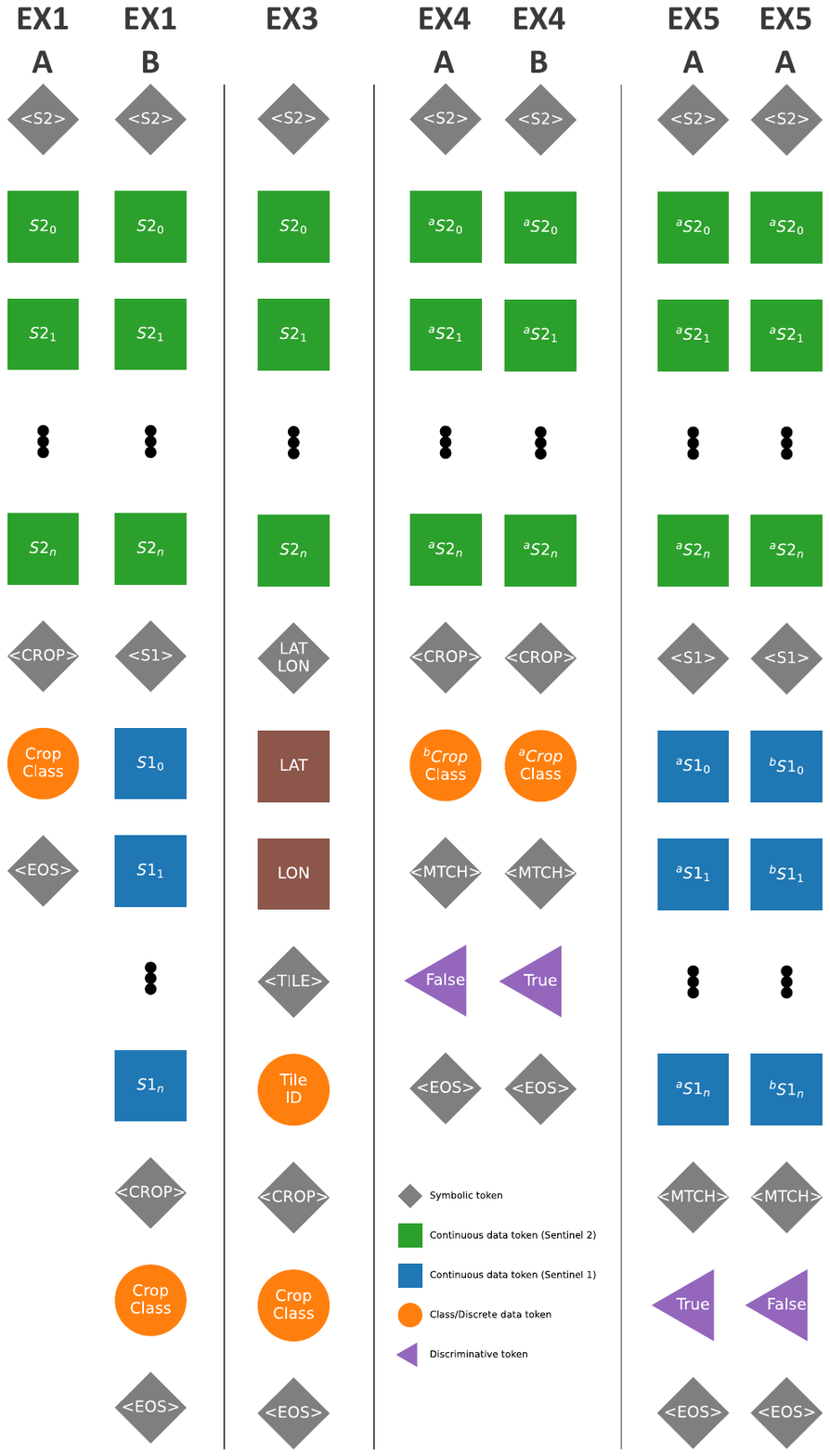}
	\caption{Symbolic illustrations of the token sequence structures used for the different experiments. Note, experiment 2 uses the same sequence as experiment 1.}
	\label{fig:fig1}
\end{figure}

We explore the capabilities of this modelling approach with the following experiments. A visualisation of the token sequences used in each experiment is shown in Figure~\ref{fig:fig1}.

\subsubsection{Experiment 1: Basic Validation}
Although LLMs and EO VLM FMs indicate we should expect transformers to be able to perform multiple classification tasks by prompting a single model with a unified token-stream framework, given the novelty of this approach and its departure from typical architectures and data and task representations for EO applications including crop type classification, we begin by presenting results showing the basic efficacy of this approach and compare to other models on the PASTIS-R dataset for both single (optical) and dual (optical + SAR) modalities. For each fold combination, a single model is trained with all the modality combinations such that it can perform any of them via prompting.

\subsubsection{Experiment 2: Comparison with EO Foundation Models}
We directly compare our model’s multi-modal crop classification performance on PASTIS-R to the PANGAEA benchmark models. PANGAEA uses a modified version of the PASTIS-R protocol using only a single fold combination (train on 1, 2, 3, validate on 4, test on 5).

\subsubsection{Experiment 3: Adding Contextual Metadata Features}
Given that crop type distributions change significantly by geographic region, using a location identifier may give the model the ability to learn regional crop priors, thus improving classification performance, particularly on easily confused crops. To test this, and test the framework's ability to easily incorporate additional data/metadata, we simply add the Sentinel tile ID and lat-lon (in decimal degrees) to the task sequences and assess the downstream impact.

\subsubsection{Experiment 4: Massively Multi-Task Modelling}
To investigate the model’s ability to perform many different tasks we construct a wide range of 29 different tasks (different combinations of inputs and outputs) from the PASTIS-R dataset and train the model to simultaneously perform all tasks in a single training run (no pre-training and fine tuning). This experiment serves as a small-scale analogue of the kind of diverse multi-task training that would characterise true promptable EO foundation models, illustrating the feasibility of that paradigm within a lightweight setup.

\subsubsection{Experiment 5: Exploiting the Multi-Task Framework and Self-Supervised Tasks to Mitigate Geographic Domain Transfer Challenges}
Geographic domain transfer is a common challenge in crop type classification. Changes between climate, seasonal patterns, management practices, crop distribution and other conditions mean a model trained on data from one geographic region often performs poorly in another region. Our framework enables very flexible combinations of tasks including both supervised and self supervised. Therefore we experiment with multi-task learning with target domain self-supervised tasks as a means of mitigating the effects of geographic domain transfer. For this experiment we train the model on supervised crop type samples from 3 of the 4 PASTIS-R Sentinel tiles, self supervised samples from the 4th tile, and evaluate crop classification performance on the 4th.

\subsection{Implementation Details and Compute}
The modelling framework is written in pytorch. All experiments were run on a desktop gaming PC running WSL with a single NVIDIA RTX 4060 Ti. Training time across experiments varied by proportion of the training dataset used, the number of epochs and sequence lengths (defined by which tasks are included in a particular run), but generally varied between 15 and 90 minutes per epoch. Benchmark training runs used 30-40 epochs without early stopping and were evaluated on the final model state. 

We optimised our model using the Adam optimiser with a batch size of 512 and employed a cosine annealing learning rate scheduler, starting from an initial learning rate of 0.001, annealing to a minimum learning rate of 0.0001 over the course of training.

\section{Experiments}

\subsection{Experiment 1 - Basic Validation}
\begin{figure}
	\centering
    \includegraphics[width=0.9\linewidth]{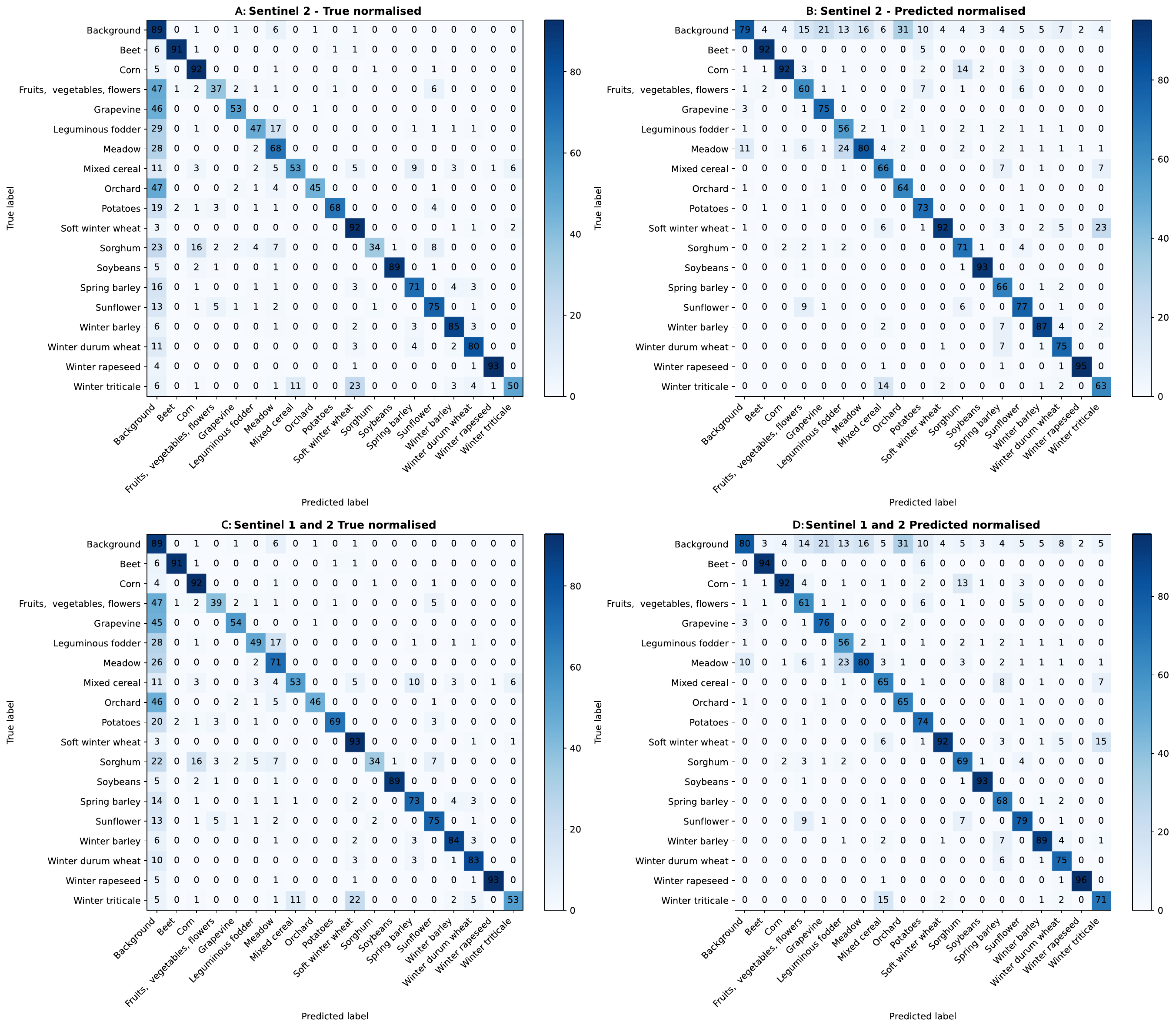}
	\caption{True (A, C) and predicted (B, D) normalised confusion matrices for Sentinel 2 only (A, B) and multi-modal Sentinel 1 + 2 (C, D) PASTIS-R crop classification on fold 5 (trained on folds 1, 2 and 3).}
	\label{fig:fig2}
\end{figure}

\begin{figure}
	\centering
    \includegraphics[width=0.6\linewidth]{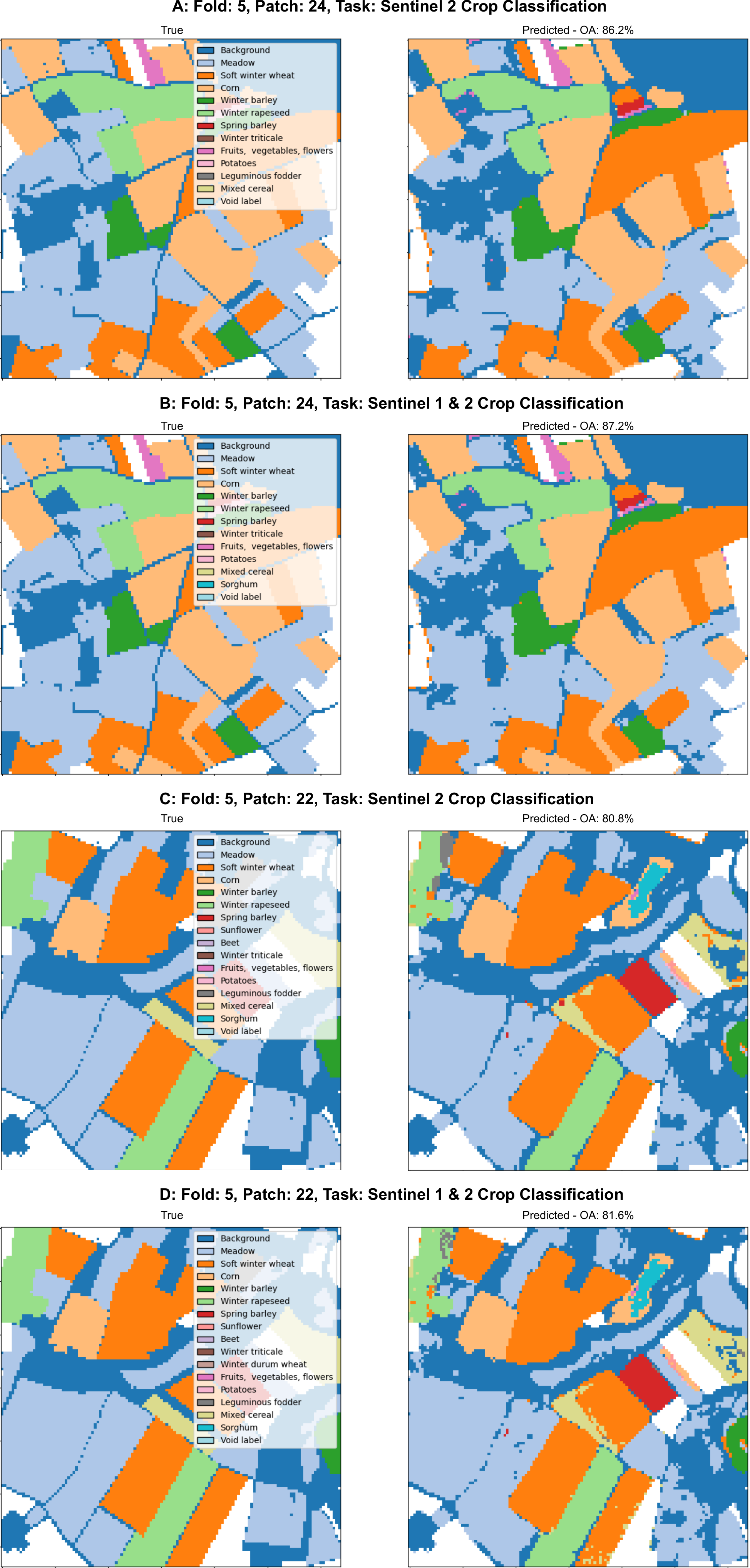}
	\caption{Inference examples (true \& predicted) for Sentinel 2 or Sentinel 1 \& 2 crop classification 2 patches in fold 5.}
	\label{fig:fig3}
\end{figure}

This section shows the key results from the basic validation experiment designed to validate the core promptable multi-task unified token-stream framing of this work. These results follow the full 5-fold cross-validation protocol of PASTIS-R and relate to the main model trained on uni- and multi-modal crop classification tasks only. Figure~\ref{fig:fig2} shows the confusion matrices for both uni-modal (Sentinel 2) and multi-modal (Sentinel 1 + 2) crop classification on the PASTIS-R dataset. The patterns of confusions (besides confusions with background) are mostly between similar crops like various cereals, between meadows and fodder, or between sorghum and corn. Figure~\ref{fig:fig3} shows example inference results indicating strong visual performance with little noise and speckle effects. Table~\ref{tab:tab1} shows our model (all modality combinations) outperform other (uni-modal) pixel-level models, while falling behind modality matched fully spatio-temporal state-of-the-art (SotA) models like U-TAE and TSViT  by roughly 5\% mIoU. In line with the fully spatio-temporal SotA models, we also find that adding Sentinel 1 data increases performance by only a small margin of 1.5-3\% mIoU.

\begin{table}[ht!]
\centering
\caption{PASTIS-R classification results of our work compared to fully spatial and other pixel level models. "-" indicates "not reported".}
\label{tab:tab1}
\begin{tabular}{llccccc}
\toprule
 & \textbf{Model} & \textbf{\# Params (×1000)} & \textbf{OA} & \textbf{mIoU} & \textbf{MA} & \textbf{Source} \\
\midrule
\multirow{3}{*}{Fully spatial} 
 & U-TAE -- S2 & 1,087 & 83.2 & 63.1 & 73.6 & \cite{garnot2021panoptic} \\
 & TSViT -- S2 & 1,700 & 83.4 & 65.4 & -- & \cite{tsvit2023} \\
 & U-TAE -- S1 + S2 (late fusion) & 1,700 & 84.2 & 66.3 & -- & \cite{garnot2021_multimodal_temporal_attention} \\
\midrule
\multirow{5}{*}{Pixel-level} 
 & MLP + LTAE -- S2 & 320 & 80.6 & -- & 65.9 & \cite{vincent2023_agri_itsc} \\
 & DTI-TS: NCC (best) -- S2 & 423 & 73.7 & -- & 59.1 & \cite{vincent2023_agri_itsc} \\
 & \textit{Ours -- S2} & \textit{465} & \textit{80.6} & \textit{59.9} & \textit{70.3} &  \\
 & \textit{Ours -- S2 + S1} & \textit{465} & \textit{81.2} & \textit{61.4} & \textit{71.6} &  \\
 & \textit{Ours -- S2 + S1 + tile + lat-lon} & \textit{465} & \textit{81.4} & \textit{62.5} & \textit{72.4} &  \\
\bottomrule
\end{tabular}
\end{table}

\subsection{Experiment 2 - Comparison with EO Foundation Models}

\begin{table}[h!]
\centering
\caption{Comparison of EO FMs on PASTIS-R as reported in \cite{pangaea2024} and \cite{terramind2025} compared to our model. Ours in italics, top in bold, best FM underlined.}
\label{tab:tab2}
\begin{tabular}{l c l}
\toprule
\textbf{Model} & \textbf{mIoU} & \textbf{Source} \\
\midrule
CROMA & 32.32 & \cite{croma2023} \\
DOFA & 30.02 & \cite{dofa2024} \\
GFM-Swin & 21.24 & \cite{gfm_swin_iccv2023} \\
Prithvi 1.0 (100M) & 33.93 & \cite{prithvi2023} \\
RemoteCLIP & 18.24 & \cite{remoteclip2023} \\
SatlasNet & 17.51 & \cite{satlaspretrain2022} \\
Scale-MAE & 24.55 & \cite{scale_mae2022} \\
SpectralGPT & 35.44 & \cite{spectralgpt2023} \\
S.-S12-MoCo & 34.49 & \cite{he2020_moco} \\
S.-S12-DINO & 36.18 & \cite{caron2021_dino} \\
S.-S12-MAE & 32.03 & \cite{he2022_mae} \\
S.-S12-Data2Vec & 34.32 & \cite{baevski2022_data2vec} \\
TerraMindv1-L & \underline{43.13} & \cite{terramind2025} \\
U-Net baseline & 31.60 & \cite{pangaea2024} \\
ViT baseline & 38.53 & \cite{pangaea2024} \\
\midrule
\textit{Our work} (SAR, Sentinel-1 only) & \textit{49.53} &  \\
\textit{Our work} (Optical, Sentinel-2 only) & \textit{58.79} &  \\
\textit{Our work} (Optical + SAR) & \textit{59.91} &  \\
\textit{Our work} (Optical + SAR + spatial metadata) & \textit{\textbf{60.91}} &  \\
\bottomrule
\end{tabular}
\end{table}

\begin{figure}
	\centering
    \includegraphics[width=0.8\linewidth]{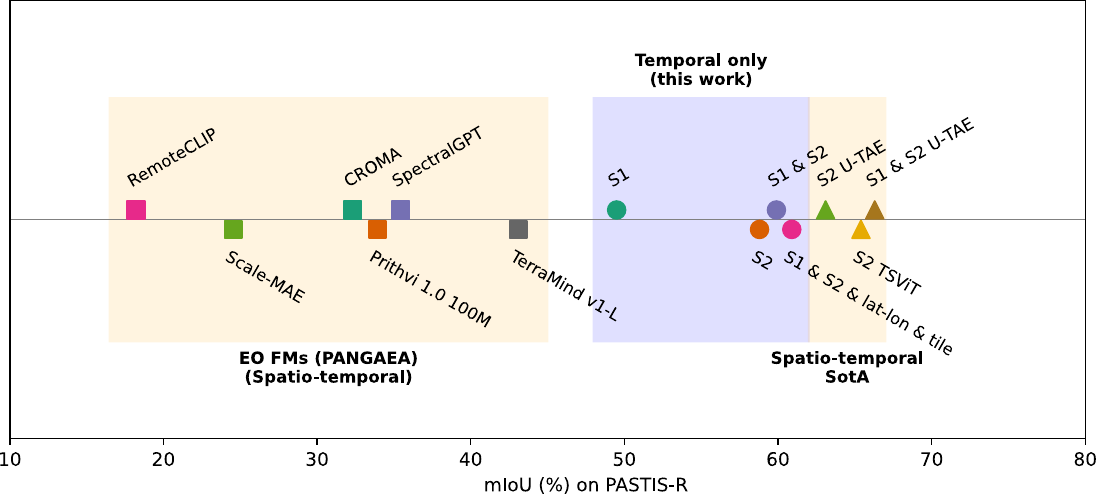}
	\caption{Illustration of the gap between spatial first EO foundation models on PASTIS-R compared with this work and comparable SotA models. References in Table~\ref{tab:tab2}.}
	\label{fig:fig4}
\end{figure}

The results in Table~\ref{tab:tab2} show our model’s performance compared to those in the PANGAEA benchmark \cite{pangaea2024}, including TerraMind \cite{terramind2025}. Note that the PANGAEA implementation of the PASTIS-R protocol uses only one fold combination (train on 1–3, validate on 4, test on 5) rather than the full 5-fold protocol used in Experiment 1. Despite using no spatial context, a single SITS-DECO model trained across all modality combinations (and prompted for each at inference) achieves substantially higher mIoU scores than all EO foundation models evaluated in PANGAEA, including those using post-model temporal encoders (L-TAE). The gap of over 15 mIoU (Figure~\ref{fig:fig4}) from the highest EO FM (TerraMind) to any SITS-DECO configuration including optical data highlights a persistent weakness in current spatial-first paradigms, which we examine further in Section 6.

\subsection{Experiment 3 - Adding Contextual Metadata Features}

\begin{figure}
	\centering
    \includegraphics[width=1.0\linewidth]{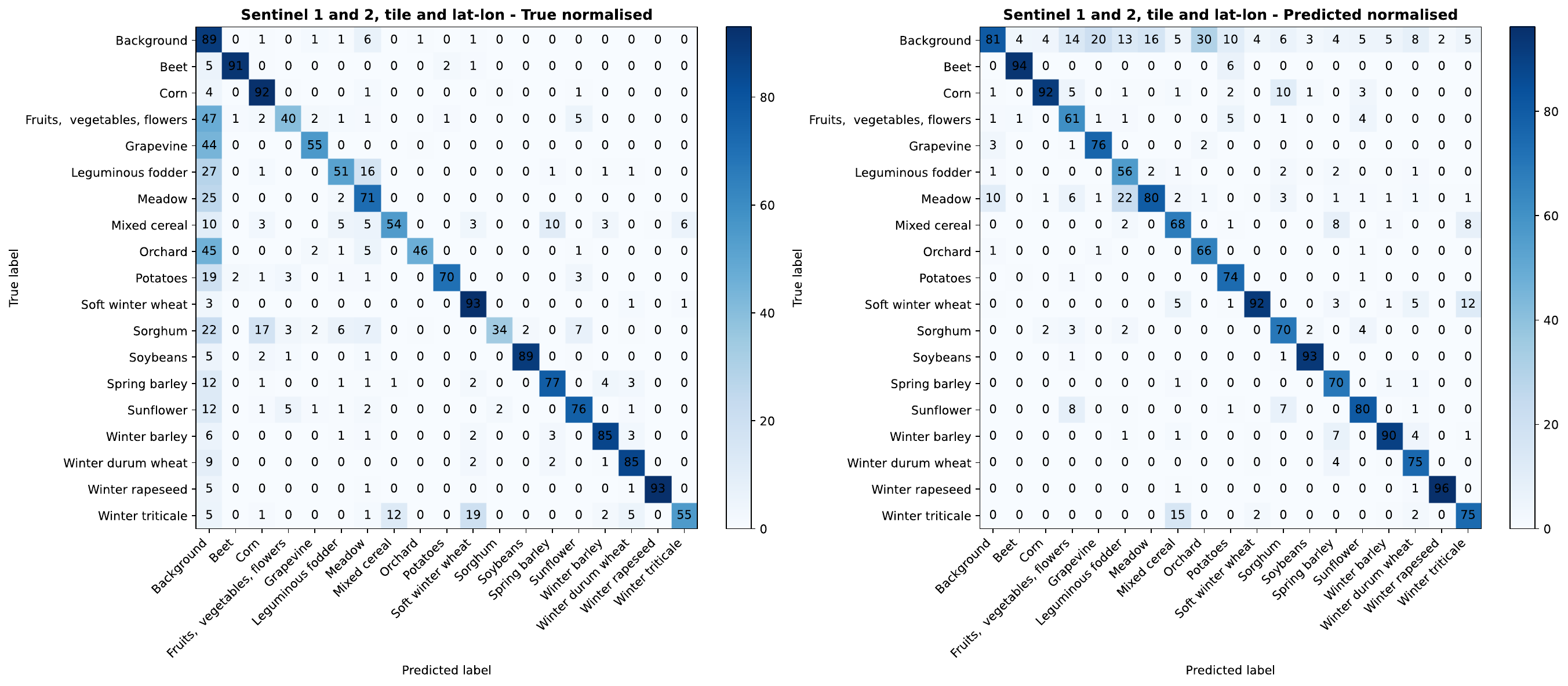}
	\caption{True and predicted normalised confusion matrices for Sentinel 1 + 2 + tile + lat-lon.
}
	\label{fig:fig5}
\end{figure}

\begin{figure}
	\centering
    \includegraphics[width=1.0\linewidth]{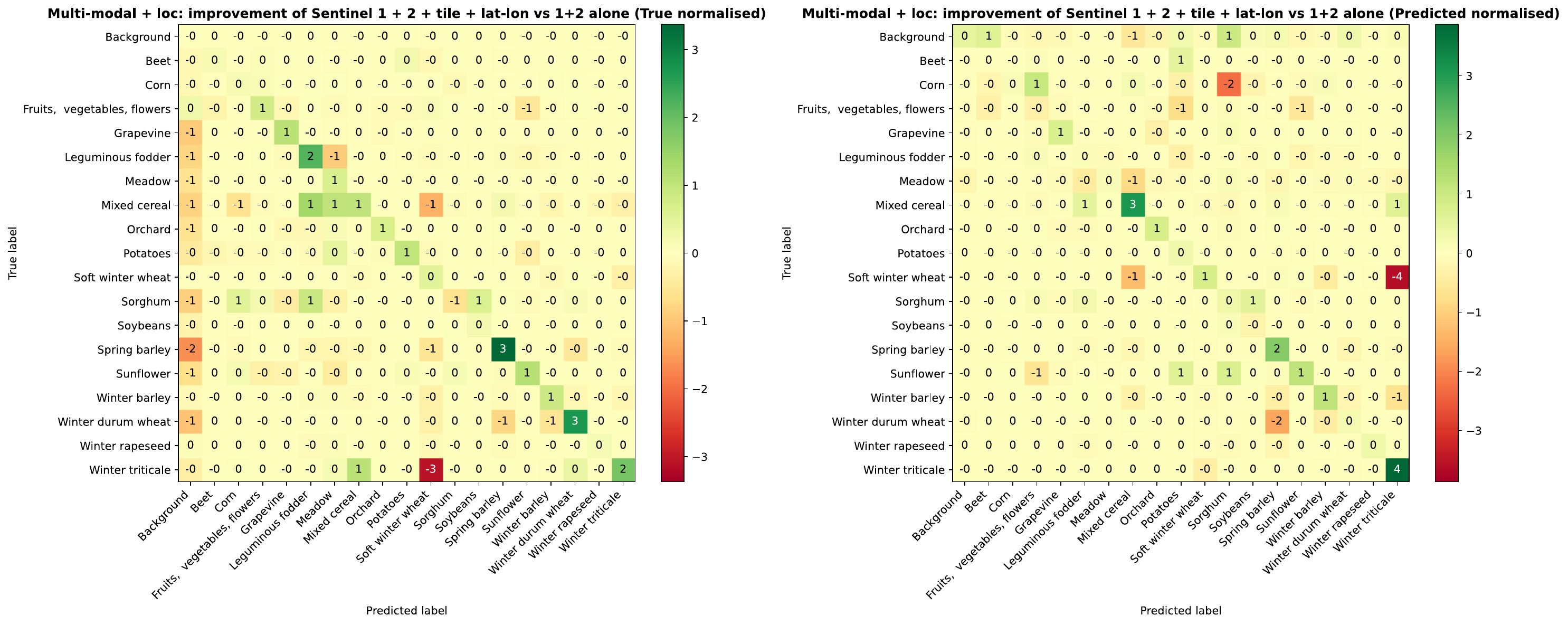}
	\caption{Relative confusion matrix showing delta/improvement over S1+2 alone. Green = increase, red = decrease. Green = diagonal means improvement.
}
	\label{fig:fig6}
\end{figure}

This experiment shows the impact of adding location metadata as model inputs in the token sequence in the form of the Sentinel 2 tile ID (categorical tokens) and the latitude and longitude (continuous tokens). Table~\ref{tab:tab1} shows the impact is modest at only 1.1\% mIoU, but Figure~\ref{fig:fig5} and ~\ref{fig:fig6} show the pattern of what has improved. The gains are mostly in easily confused classes where there are clear geographic patterns in occurrence, particularly concerning the various cereal classes (wheat, barley, triticale). 

\subsection{Experiment 4 - Massively Multi-Task Modelling}

\begin{table}[h!]
\centering
\caption{
Overall accuracy results for the massively multi-task experiment. 
\textit{*} Task includes multiple categorical targets; accuracy considers all (e.g., tile \& crop), not only the final task. 
\textit{**} Trivial for the model to memorise based on the data structure itself.
}
\label{tab:tab3}
\begin{tabular}{l c}
\toprule
\textbf{Task} & \textbf{Overall Accuracy (\%)} \\
\midrule
S2 $\rightarrow$ Crop & 80.3 \\
S1 $\rightarrow$ Crop & 77.3 \\
S2 + S1 $\rightarrow$ Crop & 79.5 \\
S2 $\rightarrow$ Tile $\rightarrow$ Crop & 80.8\textsuperscript{*} \\
S2 + latlon $\rightarrow$ Tile $\rightarrow$ Crop & 80.4\textsuperscript{*} \\
S2 + latlon $\rightarrow$ Crop & 80.7 \\
S1 $\rightarrow$ Tile $\rightarrow$ Crop & 77.4\textsuperscript{*} \\
S1 + latlon $\rightarrow$ Tile $\rightarrow$ Crop & 77.4\textsuperscript{*} \\
S1 + latlon $\rightarrow$ Crop & 77.4 \\
S2 + S1 $\rightarrow$ Tile $\rightarrow$ Crop & 79.6\textsuperscript{*} \\
S2 + S1 + latlon $\rightarrow$ Tile $\rightarrow$ Crop & 79.5\textsuperscript{*} \\
S2 + S1 + latlon $\rightarrow$ Crop & 79.6 \\
S2 $\rightarrow$ Tile & 100\textsuperscript{**} \\
S1 $\rightarrow$ Tile & 100\textsuperscript{**} \\
S2 + S1 $\rightarrow$ Tile & 100\textsuperscript{**} \\
S2 $\leftrightarrow$ S1 discrimination & 85.3 \\
S1 $\leftrightarrow$ crop discrimination & 83.8 \\
S2 $\leftrightarrow$ crop discrimination & 86.4 \\
S2 + S1 $\leftrightarrow$ crop discrimination & 85.2 \\
S1 $\leftrightarrow$ tile discrimination & 100\textsuperscript{**} \\
S2 $\leftrightarrow$ tile discrimination & 100\textsuperscript{**} \\
S2 + S1 $\leftrightarrow$ tile discrimination & 100\textsuperscript{**} \\
Crop $\rightarrow$ Tile & 39.5 \\
S2 $\rightarrow$ Crop $\rightarrow$ Tile & 66.9\textsuperscript{*} \\
S1 $\rightarrow$ Crop $\rightarrow$ Tile & 69.6\textsuperscript{*} \\
S2 + S1 $\rightarrow$ Crop $\rightarrow$ Tile & 78.3\textsuperscript{*} \\
S1 + latlon $\rightarrow$ Tile & 100\textsuperscript{**} \\
S2 + latlon $\rightarrow$ Tile & 100\textsuperscript{**} \\
S2 + S1 + latlon $\rightarrow$ Tile & 100\textsuperscript{**} \\
\bottomrule
\end{tabular}
\end{table}

This experiment is designed to test the multi-task modelling ability of our framework by creating as many possible task structures as possible from the PASTIS-R dataset and training a single model to be able to achieve them all through input prompting/token sequence structuring alone. The model trained to do this is architecturally identical to that used in experiments 1-3 except that the training data is structured into additional token sequence structures. This experiment was also trained on a different fold combination (train on 1-4, test on 5 - no validation/early stopping) than the PASTIS-R protocol so the results are non-comparable and are proof of concept only. Table~\ref{tab:tab3} shows that the model has skill on all the tasks. Some tasks are trivial for the model to memorise. Of particular note is the performance on the discrimination tasks - for all modality combinations, crop discrimination performance (does the crop label provided match the observational data?) outperform the direct crop classification tasks. Further, the performance patterns seen in experiments 1 and 3 concerning improved performance with multiple modalities are not clearly seen here or are inverse: Sentinel 1 + 2 underperforms Sentinel 2 alone.

\subsection{Experiment 5 - Geographic Domain Transfer}

\begin{table}[h!]
\centering
\caption{
Crop classification mIoU on target tile 32ULU with and without self-supervised learning (SSL). 
Values in parentheses indicate the improvement from SSL pre-training.
}
\label{tab:tab4}
\begin{tabular}{l c c c}
\toprule
 & \textbf{S2} & \textbf{S2 + Tile ID} & \textbf{S1 + S2 + Tile ID} \\
\midrule
No SSL & 27.9 & 27.6 & 30.8 \\
With SSL & \textbf{38.3} (+10.4) & \textbf{39.1} (+11.5) & \textbf{40.1} (+9.3) \\
\bottomrule
\end{tabular}
\end{table}

\begin{figure}
	\centering
    \includegraphics[width=0.5\linewidth]{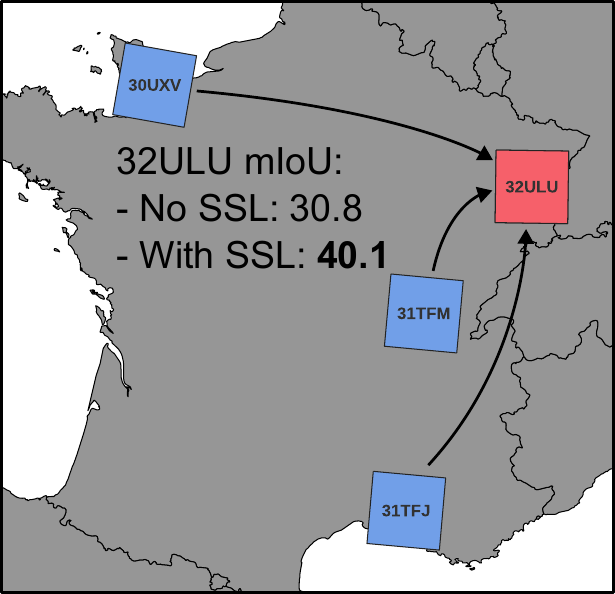}
	\caption{Map showing the source and target tiles for the SSL experiment.
}
	\label{fig:fig7}
\end{figure}

This experiment leverages the flexibility of the token sequences to apply a self-supervised geographic domain transfer mitigation. A pair of models were trained from scratch with all multi-modal crop classification tasks with data only from tiles 30UXV, 31TFM and 31TFJ (Figure~\ref{fig:fig7}), with one of the models also trained with two additional tasks on all tiles including the 32ULU target tile: 1. self-supervised discrimination task (distinguish if paired Sentinel 2 and Sentinel 1 time series come from the same pixel or not), and 2. Predict tile based on each multi-modal input data combination (Sentinel 2 and/or Sentinel 1). This experiment was not performed with the full PASTIS-R protocol of 5-fold cross validation, instead using the same protocol used in experiment 4). The results (Table~\ref{tab:tab4}) show roughly a 10\% mIoU increase on all sensor/metadata combinations. However, performance is still significantly lower than fully supervised equivalents by about 20\% mIoU.

\section{Discussion}

\subsection{Validation of Unified Token Sequence Approach}
This work set out to test a central hypothesis: that Earth Observation (EO) tasks can be framed and performed as unified token sequences within a simple, decoder-only transformer—mirroring the generative, next-token prediction paradigm that underpins large language models. The aim was not to optimise performance through architectural complexity, but to validate whether this framing itself is viable and practically useful for EO data.

Our results confirm that this sequence-based generative framing works. The model successfully performs EO classification tasks in a purely decoder-only configuration, predicting crop types directly as continuations of token streams. This demonstrates that the GPT-style unified generative token sequence formulation, traditionally applied to language or multi-modal text–image domains, extends effectively to dense, multi-temporal EO data. The approach proves that transformer decoders can model multi-temporal dependencies in EO data without requiring separate temporal encoders or modality-specific components.

Notably, all modality combinations (e.g., Sentinel-2 only, Sentinel-1 + 2, and variants including metadata) are handled by a single unified model. Each configuration is elicited through the structure of the input sequence including the data tokens and symbolic task tokens, not by training separate models. This establishes that EO classification and related tasks can be unified within one decoder-only model and performed on demand simply by altering the token sequence—an important practical and conceptual step towards promptable EO modelling. This ability to perform multiple tasks through symbolic prompting reflects the model’s inference-time flexibility. Later sections show that the same token-sequence mechanism also allows training-time extensibility, where new tasks can be introduced purely through data without modifying the architecture.

A second key finding concerns how multiple modalities are represented within the token sequence itself. Traditional EO architectures often process each modality separately and fuse features later, or temporally align modalities and concatenate them in the channel dimension. More recent work, such as Galileo \cite{galileo2025}, showed that concatenating modalities directly in the sequence dimension, so that the model attends jointly across time and modality, can simplify multi-modal learning. SITS-DECO confirms and extends this principle in a decoder-only, generative setting. We concatenate all modalities directly as sequential tokens and allow the transformer to infer their relationships implicitly through attention. The success of this simple formulation shows that effective multi-modal reasoning can emerge purely from sequence structure, reinforcing a central theme of this work: capability arises from data representation and token organisation rather than from architectural complexity.

A further insight from our results relates to what TerraMind \cite{terramind2025} describes as “thinking in modalities”, the ability of a generative model to reason across data types by performing intermediate tasks. While TerraMind achieves this through an encoder–decoder architecture, SITS-DECO demonstrates a decoder-only analogue: the ability to perform multiple related tasks within a single token stream. Because each symbolic task token defines the type of the following tokens, the model can execute chained subtasks such as predicting tile ID before crop type within one continuous generative sequence. This shows that the model can internally transition between modalities or subtasks through symbolic cues alone, without requiring separate encoders or decoders. In this sense, SITS-DECO achieves a lightweight form of “thinking in modalities” entirely within a unified generative sequence.

The effectiveness of this minimal configuration highlights an important design principle: architectural simplicity is not a limitation but a deliberate outcome of a data-centric philosophy. By encoding task and modality structure directly into the token sequence, we shift complexity from the network design to the data representation itself. This separation of concerns, in which the architecture remains simple while the data representation carries expressiveness, enables generality, reusability, and straightforward extension of the same model to new modalities or tasks without structural change.

Together, these results validate the central paradigm of this work: that a simple decoder-only, next-token prediction framework, trained on continuous and symbolic token sequences, can model complex EO data and tasks. The success of this minimalist approach shows that a generative, token-based view of EO modelling is both viable and promising, providing a conceptual bridge toward scalable, fully generative EO foundation models.

\subsection{Limitations of Current EO FMs in Multi-Temporal Settings}
The large performance gap between SITS-DECO and existing EO foundation models on PASTIS-R is striking but not anomalous: similar disparities are observed between spatial-first EO FMs and the true spatio-temporal SotA. This pattern, evident despite the use of post-model temporal encoders such as L-TAE in the PANGAEA benchmark, suggests that much of the fine-grained temporal signal is lost during spatial embedding and cannot be recovered by subsequent temporal aggregation. Our findings reinforce that dense, irregular multi-temporal information remains under-represented in current EO FM paradigms. For tasks such as crop classification, where temporal dynamics dominate the predictive signal, explicit temporal modelling provides larger gains than further spatial refinement. The fact that lightweight, pixel-based models, using raw time series can outperform foundation-scale architectures underscores the need to re-evaluate how temporal information is incorporated in EO FM design. Future work should investigate this systematically to understand and close the temporal-representation gap in current models. While this multi-temporal finding is important in its own right, it represents an empirical insight complementary to, rather than defining of, the token-sequence paradigm proposed in this work.

\subsection{Promptability Without Language}
While our approach is motivated by the path to full language integration, the use of symbolic tokens allows promptability without the significant computational and data overhead of using full language. The cost however is that while SITS-DECO can perform multiple tasks, it cannot do open-ended or zero shot tasks (i.e. tasks it hasn’t explicitly seen before). This places SITS-DECO as an instruction-centric model but lighter weight than language based instruction-centric models. Full language promptability and generation require moving fully into the GPT paradigm and discretising the EO data itself, for example via a VQ-VAE approach \cite{vqvae2017}.

\subsection{Metadata}
Our approach to adding contextual location metadata in the form of lat-lon and tile ID appended to the input sequence is not sophisticated, yet delivers modest but real and structurally expected improvements. The improvement in difficult to distinguish crops (e.g. small grains) which have spatial distribution differences seems to indicate the model is capable of learning useful distribution priors from the location metadata. The value of this should be contrasted with dedicated architectures (e.g. \cite{barriere2022_metadata}) designed to handle such metadata. Further, we did not investigate the differential impacts of lat-lon and tile ID, and there are numerous additional representations which could be used instead (e.g. geographic location encoding: \cite{russwurm2023_geoenc}). Future work should perform a thorough ablation to better understand the effect of different representations. Furthermore, adding metadata in this way may increase overfitting and reduce ability to transfer the model to unseen regions, though also provides a route to a kind of self supervised training task (see domain adaptation below). Therefore, the use of this approach is situation dependent.

\subsection{Multi-Task Modelling}
The capability of the model to perform 29 different task combinations by prompting alone (even if they are all based on a single dataset) indicates the model has learned diverse representations to allow it to generate different tokens in various different contexts. This is analogous to how LLMs function, though at a far smaller scale. In theory, additional tasks can be added simply by constructing new token sequences. New tokens (e.g. symbolic task tokens or categorical tokens) can be added by simply increasing the model input dimensionality.

However, we do observe unexpected inconsistencies in performance between tasks. Sentinel 1+2 crop classification performs slightly worse than Sentinel 2 alone (the opposite of the pattern expected and seen in experiment 1). This is likely a task interference or model capacity issue. We would suggest experimenting with larger models and loss weighting, along with the frequency of different tasks in the training set.

\subsection{Regression Tasks}
Regression tasks are potentially valuable in downstream applications. Imputing cloud gaps in Sentinel 2 data using Sentinel 1 as context is useful for data cleaning, and forecasting both Sentinel 1 and Sentinel 2 data into the future is valuable for applications like crop yield forecasting. Further, inter-sensor translation is potentially a valuable pre-training task. The viability of such regression tasks is established by EarthPT \cite{earthpt2023}. However, while the SITS-DECO code is built to enable regression tasks, such tasks were not the principal focus of this work. The authors ran initial experiments but failed to get skilful results and due to time and resource constraints were unable to further investigate training with regression tasks.

For the benefit of future work we here discuss our initial diagnoses using EarthPT as a comparison. While we followed a similar model architecture to EarthPT as explained in the methods section, several key differences in our approach compared to EarthPT remain: 1. EarthPT uses as both input and target, a single modality without clouds and with consistent cadence (the proprietary “ClearSky” product - Sentinel 2 data with cloud gaps imputed via a deep learning algorithm using Sentinel 1 data). 2. The model is fed with tokens corresponding to the date of the current and following observation, providing a “hint” to the model for the time step to predict. 3. Our model is significantly smaller than EarthPT by a factor of 100-1,000 (ours: 460k, EarthPT: 10M to 700M). We believe that it is very likely a combination of these factors (multiple modalities, cloud noise, non-fixed cadence, lack of temporal hints, model size) caused a lack of promising initial results for SITS-DECO with regression tasks and future work should address those challenges to expand both the richness of pre-training tasks and downstream applications SITS-DECO is capable of performing.

\subsection{Minimal Pre-Processing and Robustness}
We deliberately chose to reduce pre-processing burden and rely on the dynamic capabilities of the transformer itself to handle challenges typical in EO time series modelling. In spite of not applying cloud masks, using irregularly spaced data (not compositing or interpolating), using simple scaling instead of normalisation and using a very simple temporal alignment between Sentinel 1 ascending and descending data, we see excellent performance on PASTIS-R. Avoiding such pre-processing or using simple methods reduces the burden of preparing data for models in production settings. However, the topic is not deeply explored in the literature and we did not perform a detailed ablation to understand the precise effects. Anecdotally the author’s suspect that these pre-processing choices may have implications for model transferability and generalisation and as such should be systematically explored in such contexts.

\subsection{Domain Transfer Via Self-Supervision}
Inter-geography or time-frame domain transfer is a ubiquitous problem in many EO applications, especially crop classification. Our experiment here was significantly time and resource constrained and intended as a small proof of concept rather than a well explored evaluation of domain transfer mitigations in our framework. However, experiment 5 nevertheless shows a promising opportunity. Our flexible multi-task framework allows numerous self supervised tasks to be expressed in data alone and used together without adapting the architecture. In contrast to many classic studies which compare individual self supervised learning tasks for domain adaptation, our framework’s ability to straightforwardly combine multiple methods opens the door to richer and more nuanced use of self-supervised learning where different methods compliment each other. This may be analogous to causal language modelling in natural language processing where many different language tasks are implicitly used simultaneously due to the diversity of the training data, all via the next token prediction objective.

On the other hand, our framework only supports tasks which can be expressed via teacher forcing (next token prediction) and currently only discrete tokens (though we hope future work will address regression targets and/or discretisation of the EO data to allow a wider range of reconstruction tasks including e.g. modality reconstruction). This means that contrastive tasks cannot be supported, although as we’ve shown, conceptually similar but not identical discrimination tasks are supported.

The uplift vs the baseline in experiment 5 is promising, though performance still lags the target domain trained model significantly. We suggest further exploring additional combinations of self supervised tasks including those not currently supported due to the discrete token limitation, and testing transferability more robustly with full cross-region hold out validation.

\subsection{Comparisons to Other Models}
Here we compare SITS-DECO to a number of other related models:
\begin{itemize}
    \item EarthPT \cite{earthpt2023}, architectural cousin:
    \begin{itemize}
        \item Similar: decoder only, architecture itself, pixel time series, generative.
        \item Different, SITS-DECO: multi-modal (not requiring proprietary fused data), classification tasks/diverse tasks (not regression only), promptable, FAR smaller, irregular cadence, minimal pre-processing
    \end{itemize}
    \item Terramind \cite{terramind2025}, multi-modal, multi-task, promptable cousin:
    \begin{itemize}
        \item Similar: multi-modal, multi-task (but fixed not open), promptable, generative
        \item Different, SITS-DECO: smaller, pixel time series (not image patches), no language (yet/smaller), flexible - no specialised encoders or decoders: extensible via data not architecture, unified input and output spaces and representations
    \end{itemize}
    \item Galileo \cite{galileo2025}, multi-modal pre-trained backbone:
    \begin{itemize}
        \item Similar: multi-modal, early fusion/sequence dimension modality concatenation
        \item Different, SITS-DECO: truly dense time series (as opposed to monthly composites), direct promptable task focused (instead of SSL focused as backbone for fine-tuning)
    \end{itemize}
    \item Tessera \cite{tessera2025}, embedding focused cousin
    \begin{itemize}
        \item Similar: multi-modal, dense multi-temporal data
        \item Different, SITS-DECO: directly promptable and task focused, multiple and ancillary self-supervision tasks
    \end{itemize}
\end{itemize}

\subsection{Efficiency and Deployment}
SITS-DECO, being a pixel time-series only model, is inherently less computationally intensive than fully spatial models like TerraMind, although it suffers the limitations of lacking spatial context. On the other hand, using full GPT style cross-attention makes the decoder more computationally intensive compared to lightweight temporal encoders like L-TAE. We trade compute for flexibility and multi-task capability. Given that the framework is more computationally intensive than L-TAE which performs similarly at the core task of crop classification, SITS-DECO shouldn’t be seen as a drop-in replacement for other model architectures on constrained problems. Rather it should be seen and used in light of its generality which allows, as discussed above, various innovative approaches like multi-tasking and domain adaptation strategies. The full code will be released on github.

\subsection{Scope and Limitations}
The objective of this work was to validate the core idea of representing EO data tasks as sequences of tokens and using a simple, generic decoder-only architecture. To keep the scope achievable, a number of specific limitations were planned:
\begin{itemize}
    \item No spatial context
    \item No language integration
    \item No discretisation of continuous data
    \item A single dataset and task family (PASTIS-R)
    \item Small model size
    \item Limited experimental ablations
    \item Limited hyper parameter exploration
\end{itemize}

\subsection{A Data Driven Pre-Training Paradigm}
The flexible capability of SITS-DECO to train on many tasks at once opens up a new paradigm for pushing EO foundation model capability which looks a lot more like language modelling: focusing more deeply on the training data than the model architecture specifics, and scaling performance and task capability via scaling data (volume and diversity) alone. This applies regardless of the future direction(s) taken.

\section{Future Directions}
This proof of concept study suggests many possible future research and application directions. We split them into two broad categories: First, those focusing on building towards a true foundation model (along different axes) and thus requiring significant resources, and second, those focusing on specific use cases and capabilities which can be investigated with more modest resources.

Foundation modelling aligned:
\begin{enumerate}[label=\arabic*.]
    \item Towards a general purpose EO foundation model:
        \begin{enumerate}[label=\alph*)]
            \item Scale EO data
            \item Additional EO/input modalities (e.g. MODIS, climate data)
            \item Deeper exploration of SSL
            \item Scale supervised tasks
        \end{enumerate}
    \item Adding text modality:
        \begin{enumerate}[label=\alph*)]
            \item Expand vocab (symbolic → full language)
            \item Increase model capacity (n decoder blocks/layer sizes)
            \item Find/develop suitable multi-temporal captioning/instruction datasets
        \end{enumerate}
    \item Full LLM/causal language modelling integration:
        \begin{enumerate}[label=\alph*)]
            \item Discretised representation of EO data (e.g. VQ-VAE codebooks)
        \end{enumerate}
    \item Spatial context:
        \begin{enumerate}[label=\alph*)]
            \item Pixel-level context via image patch tokens (ViT style)
            \item Chip-level context via spatial EO FM chip embeddings
        \end{enumerate}
    \item Efficiency:
        \begin{enumerate}[label=\alph*)]
            \item Alternative attention mechanisms
            \item Hyperparameter exploration
        \end{enumerate}
\end{enumerate}

Exploring leveraging generative EO modelling (non-foundational):
\begin{enumerate}[label=\arabic*.]
    \item Further explore implementing geographic domain transfer mitigations including:
        \begin{enumerate}[label=\alph*)]
            \item Additional self supervised tasks (e.g. modality reconstruction)
            \item Additional auxiliary/supplementary tasks: e.g. supervision using landcover maps or other existing labels in a target region
            \item Combining multiple strategies in parallel - e.g. various self supervised and auxiliary tasks, leveraging the multi-task nature of the framework

        \end{enumerate}
    \item Further explore multi-task flexibility:
        \begin{enumerate}[label=\alph*)]
            \item Test on multiple time series components of PANGAEA (or similar) - train one model and prompt to perform the benchmark.
            \item Complimentary multi-task datasets, e.g. crop classification + planting/harvest date estimation + cover crop management + tillage management.
            \item Sequential classification, e.g. crops over multiple years or double cropping within a year (generate multiple classifications of the same task type for a given time series).

        \end{enumerate}
    \item Further explore provision of contextual data:
        \begin{enumerate}[label=\alph*)]
            \item Advanced geo-location information (beyond lat-lon) as input.
            \item Historic information as input (e.g. known cropping history).
            \item Semantic contextual data as input (e.g. history of cropping in region/other local fields).

        \end{enumerate}
\end{enumerate}

\section{Conclusions}
We show that modelling diverse EO tasks as unified sequences of tokens with a minimal decoder-only transformer is both viable and competitive with the state of the art for similar models (multi-modal pixel time series) on a single benchmark. This approach enables simple, data-driven multi-modal modelling, multi-task learning, and new strategies for domain adaptation.

It also represents a clear shift away from traditional architecture-mediated approaches toward a data-centric paradigm, akin to modern language modelling, where new capabilities emerge primarily through the addition and curation of training data. We further show that the transformer decoder can handle dense, unaligned, and noisy multi-modal time series data effectively in the context of crop classification.

The model’s ability to be prompted to perform multiple tasks without re-training opens the door to general-purpose EO modelling, mirroring the flexibility seen in language models. Although this study deliberately omits spatial context and full language integration, it demonstrates a new and extensible concept for EO modelling—with numerous paths forward, including adding language, spatial context, and large-scale pre-training toward a zero-adaptation EO foundation model.

\section{Contributors}
Samuel Barrett developed the original research concept, the initial code base and initial drafts of the manuscript and associated poster presentation.

Docko Sow contributed to development, iteration and testing of the code, and contributions to the manuscript and poster presentation.

\section{Funding Declaration}
This project was undertaken as a personal research collaboration by the two authors independently of their employers. No external funding was received and the authors volunteered their own time and computational resources. Subsequent to the experimental work, LGND AI generously provided funding to travel to the Living Planet Symposium 2025 to present this work as a poster presentation.

\section{Acknowledgements}
Samuel Barrett would like to thank Bruno Sánchez-Andrade Nuño and Konstantin Klemmer for valuable discussions on the poster presented at Living Planet Symposium 2025. He also thanks his wife for her patience and support.

\section{Generative AI Usage Statement}
This work is a product of rich multi-faceted collaboration between its human authors and Generative AI, though the human authors claim full authorship and full responsibility for the work.
AI tools used include:
\begin{itemize}
    \item ChatGPT (models: GPT-4, GPT-4o, GPT-4.5, o1, o1-pro, o3, GPT-5-instant/thinking/pro)
    \item ChatGPT Codex
    \item Github Copilot (GPT-4o, GPT-4.1, o1, o3, o3-mini)
    \item Windsurf
\end{itemize}

Generative AI was used by the authors in diverse ways, including:
\begin{itemize}
    \item Ideation and concept refinement
    \item Supporting the drafting, writing and editing manuscripts
    \item Writing and debugging code
    \item Literature search and paper summarisation

\end{itemize}

\bibliographystyle{unsrtnat}
\bibliography{references}

\end{document}